\documentclass[sigconf,natbib=true]{acmart}
\usepackage{CJKutf8} 

\usepackage{multirow}
\usepackage{arydshln} 
\usepackage{amssymb,enumitem}
\usepackage{makecell}

\def\Figref#1{Figure~\ref{#1}}
\newcommand{\tabref}[1]{Table~\ref{#1}}

\newcommand{\framework}{\textbf{Self-Selection}}
\newcommand{\approach}{\textbf{Self-Selection-RGP}}

\newcommand{\headernodot}[1]{\vspace*{1mm}\noindent\textbf{#1}}
\newcommand{\header}[1]{\headernodot{#1.}}

\AtBeginDocument{%
  }

\setcopyright{acmlicensed}
\copyrightyear{2018}
\acmYear{2018}
\acmDOI{XXXXXXX.XXXXXXX}
\acmConference[Conference acronym 'XX]{Make sure to enter the correct
  conference title from your rights confirmation emai}{June 03--05,
  2018}{Woodstock, NY}
\acmISBN{978-1-4503-XXXX-X/18/06}

\begin{document}

\title{Optimizing Knowledge Integration in Retrieval-Augmented Generation with Self-Selection}



\author{Yan Weng}
\affiliation{%
  \institution{University of Science and Technology of China}
    \city{Hefei}
    \country{China}  
    }
\email{wengyan@mail.ustc.edu.cn}

\author{Fengbin Zhu}
\authornote{Corresponding authors.}
\affiliation{%
  \institution{National University of Singapore}
  \country{Singapore}
  }
\email{zhfengbin@gmail.com}

\author{Tong Ye}
\affiliation{%
  \institution{Institute of Dataspace, Hefei Comprehensive National Science Center}
    \city{Hefei}
    \country{China} 
    }
\email{tye9601@gmail.com}

\author{Haoyan Liu}
\affiliation{%
  \institution{University of Science and Technology of China}
    \city{Hefei}
    \country{China}  
    }
\email{liuhaoyan@ustc.edu.cn}

\author{Fuli Feng}
\authornotemark[1]
\affiliation{%
  \institution{University of Science and Technology of China}
    \city{Hefei}
    \country{China}  
    }
\email{fulifeng93@gmail.com}

\author{Tat-Seng Chua}
\affiliation{%
  \institution{National University of Singapore}
  \country{Singapore}
  }
\email{dcscts@nus.edu.sg}

\renewcommand{\shortauthors}{Trovato et al.}

\begin{abstract}
Retrieval-Augmented Generation (RAG), which integrates external knowledge into Large Language Models (LLMs), has proven effective in enabling LLMs to produce more accurate and reliable responses.
However, it remains a significant challenge how to effectively integrate external retrieved knowledge with internal parametric knowledge in LLMs.
In this work, we propose a novel \framework~RAG framework, where the LLM is made to select from pairwise responses generated with internal parametric knowledge solely and with external retrieved knowledge together to achieve enhanced accuracy. 
To this end, we devise a \approach~method to enhance the capabilities of the LLM in both generating and selecting the correct answer, by training the LLM with Direct Preference Optimization (DPO) over a curated Retrieval-Generation Preference (RGP) dataset. 
Experimental results with two open-source LLMs (i.e., Llama2-13B-Chat and Mistral-7B) well demonstrate the superiority of our approach over other baseline methods on Natural Questions (NQ) and TrivialQA datasets. 

\end{abstract}



\received{20 February 2007}
\received[revised]{12 March 2009}
\received[accepted]{5 June 2009}

\maketitle

\begin{figure}[h]
  \centering
  \includegraphics[width=\linewidth]{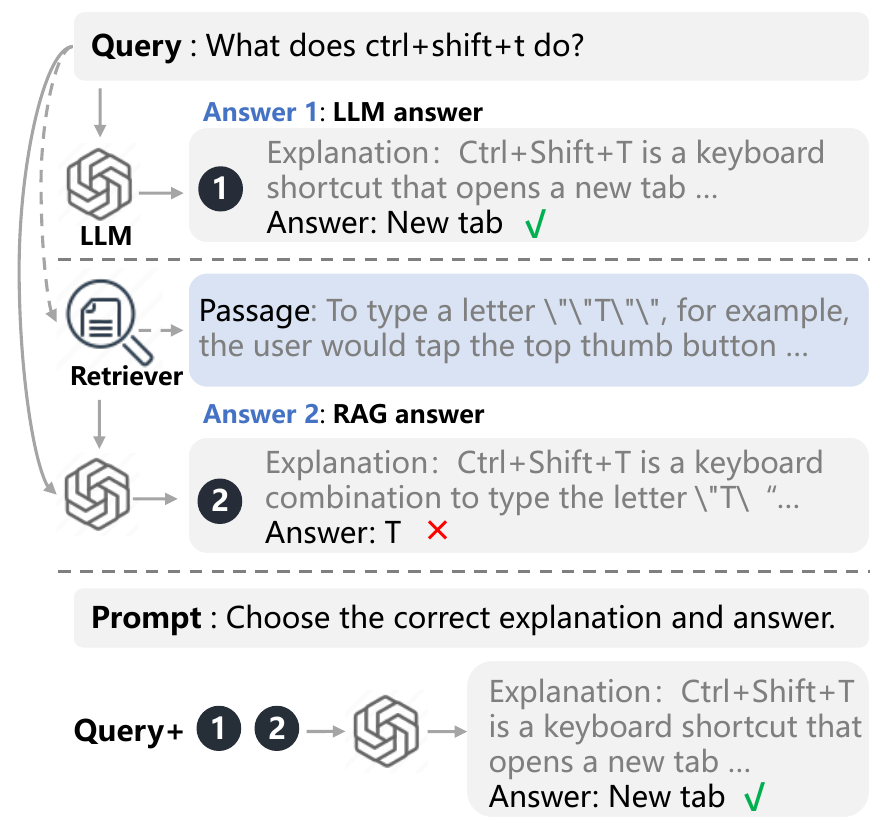}
  \caption{An illustration of the proposed \framework~framework. For a given query, 
   an LLM is first requested to generate the answers and their respective explanations 
   with and without external knowledge. 
   Then, the LLM is adopted to take them as input and choose one from them as the final answer with its explanation. }
   \label{sample}
\end{figure}

\section{INTRODUCTION}
Large Language Models (LLMs) have demonstrated remarkable capabilities across various tasks~\cite{Brown2020Few-Shot,Touvron2023LLaMA,openai2024gpt4}.
However, their reliance on static parametric knowledge~\cite{Kasai2023RealTimeQA, mallen2023trust} often leads to inaccuracy or hallucination in responses~\cite{Welleck2020Neural, min2023factscore}.
Retrieval-Augmented Generation (RAG)~\cite{Lewis2020RAG,Guu2020REALM,ram2023context,asai2023retrieval} supplements LLMs with relevant knowledge retrieved from external sources, attracting increasing research interest.
One critical challenge for existing RAG systems is how to effectively integrate internal parametric knowledge with external retrieved knowledge to generate more accurate and reliable results.

In existing RAG approaches, LLMs depend either \emph{highly} or \emph{conditionally} upon external knowledge.
The former consistently uses the retrieved content as supporting evidence~\cite{Lewis2020RAG, Guu2020REALM, trivedi2023interleaving}, which often introduces irrelevant or noisy information and overlooks the valuable internal knowledge in LLMs, resulting in sub-optimal results.
In comparison, the latter integrates external knowledge into LLMs conditionally based on specific strategies, such as characteristics of input query \cite{mallen2023trust, jeong2024adaptive, wang2023skr}, probability of generated tokens~\cite{jiang2023active, su2024dragin}, or relevance of retrieved content~\cite{zhang2023merging, Xu2024Search, liu2024raisf}.
The query-based and token-based strategies generally utilize a fixed question set or a predefined threshold to decide whether to incorporate external knowledge, limiting their effectiveness due to incomplete information; 
the relevance-based strategy employs an additional validation module to assess the retrieved content, with its accuracy largely determining the quality of the final responses.

In this work, we explore leveraging the \textbf{LLM itself} to determine the correct result by \textbf{holistically} evaluating the outputs generated with and without external knowledge.
As illustrated in \Figref{sample}, given a query ``What does Ctrl+Shift+T do?'', we instruct the LLM to generate the \emph{LLM Answer} (i.e., ``New tab'') and the corresponding explanation (i.e., reasoning steps) with its internal parametric knowledge.
Meanwhile, we employ a retriever to obtain the relevant passages from external knowledge bases and feed the query and the retrieved passages to the LLM to produce the \emph{RAG Answer} (i.e., ``T'') and the corresponding explanation. 
Next, we instruct the LLM to take the query, \emph{LLM Answer} with its explanation and \emph{RAG Answer} with its explanation as input to choose the more accurate one (i.e., ``New tab'').
In this manner, the relevant internal and external knowledge to the query is comprehensively considered, facilitating the LLM in generating accurate responses, while the RAG framework maintains its simplicity by not requiring additional modules.


Accordingly, we devise a novel \framework\ RAG framework that empowers the LLM to identify the more accurate answer to a query by evaluating both \emph{LLM Answer} and \emph{RAG Answer}, along with their respective explanations.
We validate the performance of the proposed \framework~framework with two open-sourced LLMs (see Section \ref{sec:main-result}) and find that it tends to fail in some scenarios, which we attribute to its limited capacity in distinguishing the correct answer from the incorrect one. 
To enhance the accuracy of the LLM selecting the right one among multiple responses generated from different knowledge sources, we develop a~\approach~method, leveraging Direct Preference Optimization (DPO)~\cite{Rafailov2023DPO} to fine-tune the LLM with a curated Retrieval-Generation Preference (\textbf{RGP}) dataset.
To construct this RGP dataset, we employ GPT-3.5~\cite{openai2024gpt4} to generate an \emph{LLM Answer} and an \emph{RAG Answer} for each query sampled from  WebQuestions~\cite{berant2013webq}, SQuAD 2.0~\cite{rajpurkar2018squad} and SciQ~\cite{welbl2017sciq}, and then retain only the pairs consisting of one correct answer and one incorrect answer, each accompanied by its corresponding explanation.
It consists of $3,756$ pairs of \emph{LLM Answer} and \emph{RAG answer} with their respective explanations, which we promise to release to the public to facilitate future research. 


With this dataset, we train two different LLMs, including Mistral-7B~\cite{jiang2023mistral7b} and LLaMa-2-13B-Chat~\cite{touvron2023llama2openfoundation}, and evaluate them on two widely used datasets, i.e., Natural Questions (NQ)~\cite{kwiatkowski2019natural} and TrivialQA~\cite{joshi2017triviaqa}.
It is demonstrated that our \approach~method consistently achieves high effectiveness across various retrieval settings and different LLMs, enhancing the robustness and stability of RAG systems.
Moreover, additional experiments reveal that our \approach~method not only enhances LLMs' ability to distinguish valid answers from noisy ones but also improves their answer generation capabilities.
We further validate the rationale of each design in our method through ablation studies, and conduct error case analyses to offer deeper insights into the limitations of our proposed method.

In summary, the major contributions of our paper are three-fold: 
\begin{itemize} []
\item  We introduce a novel \framework~ RAG framework that leverages LLMs to determine the correct answer by evaluating a pair of responses generated with internal parametric knowledge solely and also with external retrieved knowledge.
\item  We propose a \approach~method that applies Direct Preference Optimization (DPO) to enhance LLMs in both identifying and generating the correct answers with a curated Retrieval-Generation Preference (RGP) dataset.
\item Extensive experiments with two open-sourced LLMs achieve superior performance on two widely-used datasets, demonstrating the effectiveness of our proposed Self-Selection framework and Self-Selection-RGP method.
\end{itemize}

\begin{figure*}[h]
  \centering
  \includegraphics[width=\textwidth]{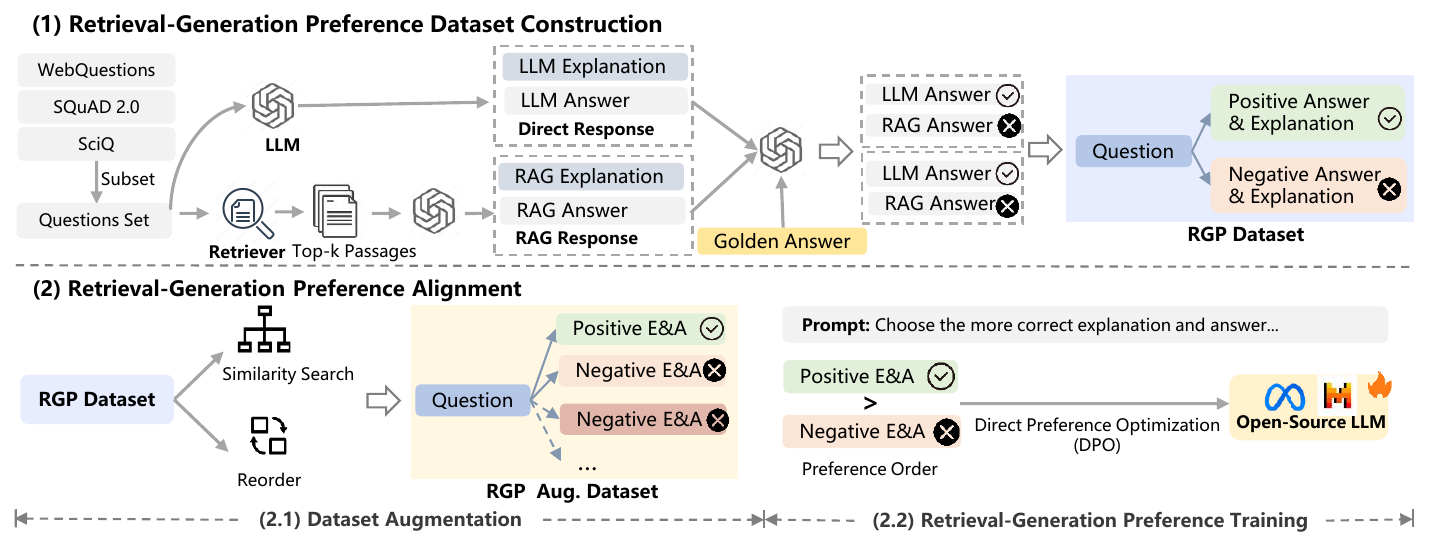}
  \caption{An illustration of the proposed \approach~method.}
  \Description{}
  \label{method}
\end{figure*}

\section{SELF-SELECTION FRAMEWORK}
In this section, we elaborate on the proposed \framework~framework for enhanced Retrieval-Augmented Generation (RAG). 
Before explaining our method, we first revisit two preliminary concepts, i.e. Large Language Model (LLM) and Retrieval-Augmented Generation (RAG).  
Then, we present the formulation of our \framework~framework with detailed notations.
In order to strengthen the capabilities of the LLM in accurately generating and selecting responses, we develop a novel \approach~method, which is essentially fining tuning the LLM over a newly built Retrieval-Generation Preference (RGP) dataset. 

\subsection{Preliminaries}
\subsubsection{\textbf{Large Language Model (LLM)}}
For an LLM represented by $\mathcal{M}$, given a prompt $\bar{p}$ and a query $q$ as inputs, it returns a textual answer $\bar{a}$ as the output, which is formally expressed as 
\begin{equation}\label{eq:llm-ans}
\bar{a} = \mathcal{M}(\bar{p}, q).
\end{equation}
\subsubsection{\textbf{Retrieval-Augmented Generation (RAG)}} 
An RAG system employs a retriever to enhance the capability of the LLM by enabling it to access external knowledge beyond its internal parametric knowledge~\cite{lee2019LatentRetrieval, Guu2020REALM}. 
Given a query $q$, the retriever $\mathcal{R}$ searches for the relevant knowledge (e.g., passages) $C$ from an external knowledge base or corpus. 
A common approach for RAG is to include the retrieved passages $C$ in the input to the LLM to improve the response quality.
Formally,
\begin{equation}
C = \mathcal{R}(q)\label{eq:retrieval},
\end{equation}
\begin{equation}\label{eq:rag-ans}
\hat{a} = \mathcal{M}(\hat{p}, q, C),
\end{equation}
where $\hat{p}$ represents the prompt used in RAG and $\hat{a}$ denotes the answer predicted by the LLM taking into account the retrieved passages $C$.

\subsection{Task Formulation}

In this part we present the formulation of our \framework~framework.
An illustration is provided in \Figref{sample}. 
Given a query $q$, we first prompt an LLM $\mathcal{M}$ with $\bar{p}$ denoting the prompt to output the answer $\bar{a}$ with its explanation $\bar{e}$, where we refer to $\bar{a}$ as the \emph{LLM Answer} and $\bar{e}$ as the \emph{LLM Explanation}.
Next, we use the retriever $\mathcal{R}$ to gather relevant passages $C$ (Eq. (\ref{eq:retrieval})) to the same query $q$.
Then, we prompt the LLM $\mathcal{M}$ with $\hat{p}$ while providing $q$ and $C$ in the input, to generate $\hat{a}$ with its explanation $\hat{e}$, where we refer to  $\hat{a}$ and $\hat{e}$  as the \emph{RAG Answer} and the \emph{RAG Explanation}.
Finally, we prompt the LLM $M$ with a prompt $p$ by taking the query $q$, the \emph{LLM Answer} $\bar{a}$ with its \emph{LLM Explanation} $\bar{e}$, the \emph{RAG Answer} $\hat{a}$ with its \emph{RAG Explanation} $\hat{e}$ as inputs to select one as the final answer $a$ and final explanation $e$.
Formally,
\begin{equation}\label{eq:llm-ans-e}
(\bar{a}, \bar{e}) = \mathcal{M}(\bar{p}, q);
\end{equation}
\begin{equation}\label{eq:rag-ans-e}
(\hat{a}, \hat{e}) = \mathcal{M}(\hat{p}, q, C);
\end{equation}
\begin{equation}\label{eq:final-ans-e}
(a, e) = \mathcal{M}(p, q, (\bar{a}, \bar{e}), (\hat{a}, \hat{e})).
\end{equation}

\subsection{Self-Selection-RGP}

\subsubsection{\textbf{Motivation}}
We evaluate the performance of the proposed \framework~framework on two widely used QA datasets, Natural Question (NQ)~\cite{kwiatkowski2019natural} and TriviaQA~\cite{joshi2017triviaqa}, using existing open-source models, including Mistral 7B~\cite{jiang2023mistral7b} and Llama2-13B-Chat~\cite{touvron2023llama2openfoundation} without any model parameter updates. 
We report the experimental results in \tabref{tab:main} of Section \ref{sec:experiments}.
We find that our \framework~framework is promising in enhancing LLMs' answer generation by fusing internal knowledge with external knowledge, but directly applying such knowledge fusion does not always bring enhancements. 
For instance, simply equipping Mistral-7B with a retriever outperforms applying our \framework~to Mistral-7B with the same retriever on the NQ dataset. 
One assumption is that LLMs struggle to reliably discern the correct answer between two candidates generated from different knowledge sources.
In essence, this knowledge selection process is consistent with the goal of preference alignment in LLMs, i.e. generating the desired (positive) sample while rejecting the undesired (negative) one from a pair of preference data.
To address this challenge, we explore tuning LLMs through preference alignment techniques to enhance their ability to discern and select the correct answer from two candidates generated by different knowledge sources.
To achieve this goal, we develop a novel \approach~method to enhance LLMs' capabilities in identifying and generating correct answers, as shown in \Figref{method}.
We first build a preference dataset, then employ a simple yet effective augmentation technique to expand it, and finally apply the augmented preference dataset to train open-sourced LLMs with Direct Preference Optimization (DPO)~\cite{Rafailov2023DPO}. 

\subsubsection{\textbf{Retrieval-Generation Preference Dataset}}
Here we explain how we build the Retrieval-Generation Preference (RGP) dataset used for fine-tuning LLMs in~\approach.

\header{Preference Candidate Generation}
We first employ an LLM to produce two sets of responses for each query $q$: (i) an \emph{LLM Answer} $\bar{a}$ with its \emph{LLM Explanation} $\bar{e}$, derived from the model’s internal parametric knowledge; and (ii) an \emph{RAG Answer} $\hat{a}$ with its \emph{RAG Explanation} $\hat{e}$, relying on the externally retrieved information.
Specifically, we randomly select a subset of QA pairs from three existing open-domain QA datasets, including WebQuestions~\cite{berant2013webq}, SQuAD2.0~\cite{rajpurkar2018squad}, and SciQ~\cite{welbl2017sciq}.
Let $\mathcal{D}$ denote the obtained set of QA pairs. Formally,
\begin{equation} \label{eq:gold}
    \mathcal{D} = \bigl\{q^{(i)}, a_g^{(i)}\bigr\}_{i=1}^N
\end{equation}
where $a_g$ is the golden answer to the query $q$, $N$ is the number of QA pairs and  $i$ is the $i$-th QA pair in $\mathcal{D}$.
For each query $q$ in $\mathcal{D}$, we utilize a retriever $\mathcal{R}$ to retrieve the top-K passages $C$ from a corpus (Eq. (\ref{eq:retrieval})).
To ensure the quality of the constructed preference dataset, we employ GPT-3.5~\cite{Ouyang2022Training} as the model $\mathcal{M}$ for candidate answer and explanation generation given a query.
According to Eq. (\ref{eq:llm-ans-e}) and Eq. (\ref{eq:rag-ans-e}), we generate the answers and explanations ($\bar{a}$, $\bar{e}$) and ($\hat{a}$, $\hat{e}$). 
Finally, we obtain a collection of preference candidates for constructing the RGP datasets.
Formally, the $D$ is expanded as
\begin{equation} \label{eq:gold}
    \mathcal{D} = \bigl\{q^{(i)}, a_g^{(i)}, \bar{a}^{(i)},\bar{e}^{(i)}, \hat{a}^{(i)}, \hat{e}^{(i)}\bigr\}_{i=1}^N.
\end{equation}

\header{Preference Data Filtering}
In the RGP dataset, each instance should include both a desired (positive) answer and an undesired (negative) answer.
We filter these required instances from the collection $\mathcal{D}$.
For each instance in $D$, we first employ GPT-3.5 to assess whether the \emph{LLM Answer} $\bar{a}$ and the \emph{RAG Answer} $\hat{a}$ are correct by comparing each to the golden answer $a_g$.
After that, we only retain the instances that contain one right answer and one wrong answer, where (i) $\bar{a}$ is correct but $\hat{a}$ is incorrect; or (ii) $\hat{a}$ is correct but $\bar{a}$ is incorrect.
Based on this strategy, we gather all appropriate instances in $\mathcal{D}$ to build our RGP dataset $\mathbb{D}$. Formally,
\begin{equation} \label{eq:gold}
    \mathbb{D} = \bigl\{q^{(j)}, a_g^{(j)}, (a_p^{(j)}, e_p^{(j)}), (a_n^{(j)}, e_n^{(j)})\bigr\}_{j=1}^M
\end{equation}
where $a_p$ and $e_p$ represent the positive answer and its explanation, $a_n$ and $e_n$ represent the negative answer and its explanation, $M$ denotes the number of instances and $j$ denotes the $j$-th instance in $\mathbb{D}$.
Finally, we retain $3,756$ preference instances in the RGP dataset.
We promise to release it for facilitating future reseach.

\subsubsection{\textbf{Retrieval-Generation Preference Alignment}}
With the constructed RGP dataset, we train open-source LLMs to enhance their ability to distinguish the positive answer from the negative counterpart.

\header{RGP Dataset Augmentation}
To improve LLMs' preference alignment, we first augment the RGP dataset through a simple yet effective approach to produce more preference instances. 
In particular, given a query $q$ in RGP, we search for the top-K similar queries in the RGP datasets and we regard all answers to these $K$ queries as negative answers to the query $q$.
Formally, for each query $q$ in the RGP dataset~$\mathbb{D}$,  we denote the obtained most similar queries and their responses in RGP as $\mathbb{G}$:  
\begin{equation} \label{eq:group}
    \small
    \mathbb{G}^{(i)} = \{ q^{(j)}, y_w^{(j)}, y_l^{(j)}~|~\underset{\text{top-}\mathrm{K}}{\mathrm{argmax}}\ S\left(q^{(i)}, q^{(j)}\right)\}~~~\forall q^{(i)} \in \mathbb{D}, i\neq j
\end{equation}
where $S(q^{(i)}, q^{(j)})$ represents the similarity score between $q^{(i)}$ and $q^{(j)}$, and $y_w^{(j)}$ and $y_l^{(j)}$ represent the corresponding positive and negative response (i.e., answer with its explanation) for $q^{(j)}$ in RGP.
For one query $q^{(i)}$, $y_w^{(i)}$ and $y_l^{(i)}$ are the original positive and negative response in RGP.
Then, we regard all $y_w^{(j)}$ and $y_l^{(j)}$ in the obtained set $\mathbb{G}^{(i)}$ as the additional negative responses to $q^{(i)}$.
For each query, now we have $1$ positive response and $2K+1$ negative responses, which can be used to form $2K+1$ pairs of preference instances. 
Let $\mathbb{D}_{aug}$ denote the augmented RGP dataset. Formally,
\begin{equation}
\mathbb{D}_{aug}^{(i)} = \left\{ \left( q^{(i)}, y_w^{(i)}, y_{lj}^{(i)} \right) \right\}_{i=1}^M, \, j=1,2,\dots,2K+1,
\end{equation}
where $y_{lj}^{(i)}$ is the $j$-th negative response to the query $q^{(i)}$. 

\header{Retrieval-Generation Preference Training} 
With the augmented preference dataset, our goal is to train open-sourced LLMs to enhance their capabilities in distinguishing the correct answers from the incorrect ones.
During the preference alignment phase of LLM training, each instance in the augmented dataset $\mathbb{D}_{aug}$ comprises three key elements: an input $x$, a desired response $y_w$, and an undesired response $y_l$,  which is denoted as $y_w \succ y_l \mid x$.
Specifically, the input $x$ consists of a query $q$, the desired response $y_w$, the undesired response $y_l$, and a prompt $p$ designed to instruct the LLM $M$ to choose between $y_w$ and $y_l$ (see Eq. (\ref{eq:final-ans-e})).
The desired response $y_w$ includes the correct answer along with its explanation, while the undesired response $y_l$ contains an incorrect answer and its explanation.
To enhance the robustness of the trained model, we randomly alternate the order of $y_w$ and $y_l$ within the input $x$.

We adopt Direct Preference Optimization (DPO)~\cite{Rafailov2023DPO} to train LLMs.
It enables preference data to be directly associated with the optimal policy, eliminating the need for any additional reward model.
DPO formulates a maximum likelihood objective as follows:
\begin{equation}\label{eq:optimum_model}
    \tiny
    \mathcal{L}_\text{DPO}(\pi_{\theta}; \pi_{ref}) = -\mathbb{E}_{(x, y_w, y_l)\sim \mathcal{D}}\left[\log \sigma \left(\beta \log \frac{\pi_{\theta}(y_w\mid x)}{\pi_{ref}(y_w\mid x)} - \beta \log \frac{\pi_{\theta}(y_l\mid x)}{\pi_{ref}(y_l\mid x)}\right)\right]
\end{equation}
where  $\beta$ represents the deviation of the policy $\pi_{\theta}$ from the reference model $\pi_{ref}$.
Our proposed optimization method aims to enhance LLMs' answer selection capability, enabling them to holistically evaluate multiple responses generated from diverse knowledge sources and identify the most accurate one among them.
In addition, we also hope this optimization method can further improve the inherent ability of LLMs in answer generation (more analysis is provided in Section \ref{sec:llm-improve}).

\section{EXPERIMENTS}
\label{sec:experiments}
In this section, we evaluate the proposed method with extensive experiments.
Note that we conduct our experiments with two variants based on our proposed \framework~framework,  i) \textbf{Self-Selection-Ori}, which refers to the RAG method that applies Self-Selection on vanilla LLMs. ii) \textbf{Self-Selection-RGP}, which denotes the RAG method that applies Self-Selection on the LLMs trained with our augmented RGP dataset.
For a comprehensive evaluation, we seek to address the following research questions:
\begin{itemize} 
\item \textbf{RQ1}: How does Self-Selection-RGP perform compared to other compared methods? 
\item \textbf{RQ2}: Whether Self-Selection-RGP is generalizable across different base LLMs and retrieval settings? 
\item \textbf{RQ3}: To what extent can Self-Selection-RGP affect the LLMs' inherent ability in answer generation? 
\item \textbf{RQ4}: What is the effect of each design in our proposed Self-Selection-RGP? 
\end{itemize}

\subsection{Experimental Setup}
\subsubsection{\textbf{Datasets}}
 We verify the proposed method with two open-domain QA datasets, Natural Question (NQ)~\cite{kwiatkowski2019natural} and TriviaQA~\cite{joshi2017triviaqa}.
\begin{itemize}[leftmargin=*,nosep]
    \item \textbf{Natural Questions}: Natural Questions is a widely used dataset for the evaluation of RAG systems. Each QA pair is annotated by human annotators based on a Wikipedia page.
    \item \textbf{TriviaQA}: TriviaQA is another popular dataset based on trivia questions, paired with independently collected evidence documents, which is designed to support challenging reading comprehension tasks.
\end{itemize}
Following prior works \cite{trivedi2023interleaving,jeong2024adaptive}, we use the same test split for each dataset with the same external corpus to evaluate RAG methods. 
We present the statistics in \tabref{tab:dataset}.  

\begin{table}[htbp]
  \centering
  \large
  \setlength\abovecaptionskip{-0.3pt}
  \setlength\belowcaptionskip{-0.3pt}
  \caption{Statistics of the test datasets.}
  \resizebox{\columnwidth}{!}{%
    \begin{tabular}{p{4cm}p{3cm}p{3cm}} 
    \toprule
    \bf Dataset & \bf \#Passages & \bf \#QA pairs \\
    \midrule
    Natural Questions (NQ) & 21,015,324 & 500 \\
    TriviaQA & 21,015,324 & 500 \\
    \bottomrule
    \end{tabular}%
  }
  \label{tab:dataset}%
\end{table}

\begin{table*}[t]
  \setlength\abovecaptionskip{-0.3pt}
  \setlength\belowcaptionskip{-0.3pt}
  \setlength{\tabcolsep}{1.9mm}
  \centering
  \caption{Main results of our proposed methods and all baseline methods. }
    \begin{tabular}{llrrrrrrrrrrrr}
    \toprule
          &       & \multicolumn{6}{c}{\textit{zero-shot}}        & \multicolumn{6}{c}{\textit{few-shot}} \\
          &       & \multicolumn{3}{c}{\textbf{NQ}} & \multicolumn{3}{c}{\textbf{TriviaQA}} & \multicolumn{3}{c}{\textbf{NQ}} & \multicolumn{3}{c}{\textbf{TriviaQA}} \\
    \multicolumn{1}{c}{Base LLM} & \multicolumn{1}{c}{Method} & \multicolumn{1}{c}{EM} & \multicolumn{1}{c}{F1} & \multicolumn{1}{c}{Acc} & \multicolumn{1}{c}{EM} & \multicolumn{1}{c}{F1} & \multicolumn{1}{c}{Acc} & \multicolumn{1}{c}{EM} & \multicolumn{1}{c}{F1} & \multicolumn{1}{c}{Acc} & \multicolumn{1}{c}{EM} & \multicolumn{1}{c}{F1} & \multicolumn{1}{c}{Acc} \\
    \midrule
    \multirow{6}[2]{*}{Mistral (7B)} & LLM Only & 21.8  & 35.5  & 34.0  & 41.2  & 53.4  & 52.6  & 20.8  & 34.9  & 32.2  & 43.8  & 54.1  & 53.4 \\
          & Standard RAG  & 35.8  & 51.2  & 51.0  & 45.8  & 58.1  & 59.8  & 37.8  & 51.6  & 50.6  & 47.2  & 59.4  & 60.4 \\
          & Self-RAG & -     & -     & -     & 29.0  & 43.2  & 60.6  & -     & -     & -     & 41.0  & 53.0  & 55.0 \\
          & SURE  & \textbf{39.0} & 52.4  & 47.6  & 48.6  & 59.7  & 61.6  & 38.8  & 51.1  & 47.6  & 49.2  & 60.1  & 61.4 \\
          & \textbf{Self-Selection-Ori} & 34.6  & 50.1  & 50.2  & 48.4  & 61.2  & 62.8  & 33.0  & 47.3  & 45.2  & 49.8  & 61.7  & 62.4 \\
          & \textbf{Self-Selection-RGP} & 37.8  & \textbf{52.5} & \textbf{53.6} & \textbf{54.4} & \textbf{66.2} & \textbf{67.0} & \textbf{40.2} & \textbf{53.8} & \textbf{53.2} & \textbf{54.4} & \textbf{66.2} & \textbf{65.4} \\
    \midrule
    \multirow{6}[2]{*}{Llama2-Chat (13B)} & LLM Only & 21.2  & 31.9  & 28.2  & 43.2  & 50.1  & 48.0  & 24.8  & 35.3  & 30.2  & 49.8  & 57.1  & 54.0 \\
          & Standard RAG  & 24.6  & 37.0  & 45.2  & 35.2  & 46.1  & 55.0  & 31.8  & 43.4  & 44.8  & 46.0  & 54.8  & 54.6 \\
          & Self-RAG & -     & -     & -     & 17.2  & 36.6  & 63.4  & -     & -     & -     & 39.0  & 52.2  & 59.0 \\
          & SURE  & \textbf{39.4} & \textbf{52.3} & \textbf{52.0} & 50.4  & 63.0  & 63.8  & \textbf{42.6} & \textbf{53.2} & 50.4  & 40.6  & 51.3  & 65.0 \\
          & \textbf{Self-Selection-Ori} & 31.6  & 43.8  & 45.2  & 43.0  & 53.5  & 56.6  & 33.2  & 44.6  & 43.2  & 49.2  & 59.9  & 60.0 \\
          & \textbf{Self-Selection-RGP} & 36.6  & 49.2  & 46.2  & \textbf{56.6} & \textbf{66.3} & \textbf{66.0} & 40.0  & 52.4  & \textbf{51.6} & \textbf{52.6} & \textbf{65.5} & \textbf{66.8} \\
    \bottomrule
    \end{tabular}%
  \label{tab:main}%
\end{table*}%
\subsubsection{\textbf{Baselines}}
We will compare our proposed methods with the following baseline methods:
\begin{itemize}[leftmargin=*,nosep]
    \item \textbf{LLM Only}: The response to each query is generated solely by LLMs. 
    \item \textbf{Standard RAG}: The response to each query is produced by LLMs after appending the retrieved passages to the input.
    \item \textbf{Self-RAG}~\cite{asai2024selfrag}: Specialized reflection tokens are utilized to enable LLMs to control retrieval and evaluate the relevance of the retrieved content during reasoning. In the experiments, we use the open-source models fine-tuned with the SELF-RAG framework, including the fine-tuned models based on Mistral 7B and Llama2-13B\cite{asai2024selfrag,SciPhi-AI2024}.
    \item \textbf{SURE}~\cite{kim2024sure}: LLMs first generate summaries of the retrieved passages for each candidate answer, and then identify the most plausible answer by evaluating each summary’s validity and ranking. We employ Mistral 7B  and Llama2-13B-Chat \cite{touvron2023llama2openfoundation} as the backbone models, consistent with our method.
\end{itemize}

\subsubsection{\textbf{Evaluation Metrics}}
In our experiments, we use the Exact Match (EM), F1 score and Accuracy (Acc) as our evaluation metrics, following the standard evaluation protocol~\cite{mallen2023trust,jeong2024adaptive,kim2024sure}. 
The EM measures whether the prediction and the gold answer are exactly the same; the F1 score measures the number of overlapping words between the prediction and the gold answer; the Accuracy metric measures whether the prediction contains the gold answer.
We normalize the predictions and gold answers (i.e., lowercasing and punctuation) to compute the metrics, following the implementation in previous works~\cite{rajpurkar-etal-2016-squad,kim2024sure}.

\subsubsection{\textbf{Implementation Details}}
In total, we sample $11,756$ QA pairs from WebQuestions~\cite{berant2013webq}, SQuAD2.0~\cite{rajpurkar2018squad}, and SciQ~\cite{welbl2017sciq}.
We use the official 2018 English Wikipedia as the external knowledge base similar to prior works~\cite{karpukhin-etal-2020-dense,jeong2024adaptive,asai2024selfrag}. 
We use pre-trained BGE (i.e., bge-large-en-v1.5)~\cite{Xiao2024cpack} as the retriever to obtain the relevant passages for each question.
For each query, we retrieve a variable number of passages, ranging from $1$ to $5$.
We adopt GPT-3.5 as our LLM for generating answers and explanations for each question, as well as evaluating the consistency between two given answers.
We utilize a Sentence transformer model (i.e., all-mpnet-base-v2)~\cite{reimers-2020-multilingual-sentence-bert} to identify the top-K similar questions for dataset augmentation.
After the data augmentation, we retain $21,928$ preference instances for model training.

For model training, we adopt the widely-used Mistral 7B (i.e., Mistral-7B-Instruct-v0.2)~\cite{jiang2023mistral7b} and Llama2-13B-Chat~\cite{touvron2023llama2openfoundation} as the base LLMs.
We apply DPO with Low-Rank Adapters (LoRA) \cite{hu2022lora} to train LLMs.
We conduct all our experiments on a GPU machine with 4 A800 NVIDIA RTX GPUs.
For model inference,  we use the same retriever BGE (i.e., bge-large-en-v1.5)~\cite{Xiao2024cpack} across all compared methods for a fair comparison.
We retrieve the top $5$ passages for each query from the external knowledge base.
Our experiments are conducted underboth zero-shot and few-shot settings. For the zero-shot setting, we follow the official settings of prior work\cite{asai2024selfrag,kim2024sure}.
For the few-shot setting, we include three examples in the prompts formatted according to the objective of each method.
Since the training data of Self-RAG includes the NQ dataset, we do not consider the results of Self-RAG on the NQ dataset in our experiments.

\begin{figure*}[t]
\setlength\abovecaptionskip{0px}
\setlength\belowcaptionskip{0px}
  \centering
  \includegraphics[width=\linewidth]{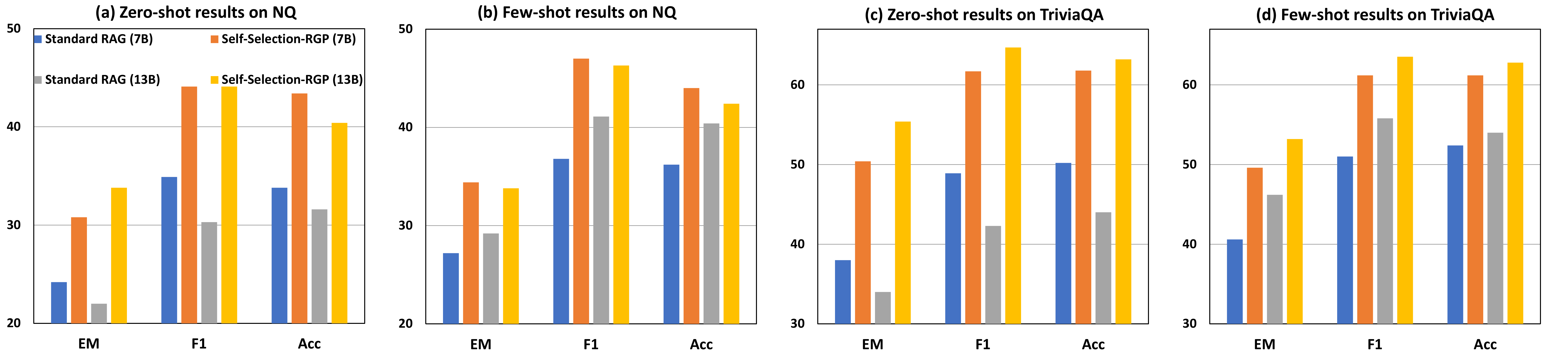}
  \caption{An illustration of the effects of using a different retriever. Our \approach~and Standard RAG both use BM25 as the new retriever. We adopt Mistral-7B as the base LLM for Standard RAG (7B) and Self-Selection-RGP (7B), and Llama2-13B-Chat for Standard RAG (13B) and Self-Selection-RGP (13B). } 
  \label{fig:retriever}
\end{figure*}
\subsection{Main Results (RQ1)}
\label{sec:main-result}

To verify the effectiveness of the proposed \framework~ framework, we compare the performance with the baseline methods.
In Table~\ref{tab:main}, we present the main results on the NQ and TriviaQA datasets with Mistral-7B~\cite{jiang2023mistral7b} and Llama2-13B-Chat~\cite{Touvron2023LLaMA}, under both zero-shot and few-shot settings. 
From the results, we make several key observations: 

\begin{itemize}[leftmargin=*,nosep]
\item With Mistral-7B as the base LLM, our Self-Selection-RGP method consistently outperforms all compared methods on NQ and TriviaQA datasets under both zero-shot and few-shot settings.
On the TriviaQA dataset, our Self-Selection-RGP model achieves an accuracy of $67.0$ in the zero-shot setting and $65.4$ in the few-shot setting, making significant improvements of $5.4$ and $4.0$ points, respectively, compared to the best baseline method SURE which scores  $61.6$ and $61.4$.
In contrast, the performance gains of our Self-Selection-RGP method on the NQ dataset are relatively smaller.
Our method scores $53.6$ and $53.2$ in accuracy under zero-shot and few-shot settings, respectively, which makes an improvement of $2.6$ and $2.6$ over the best baseline method Standard RAG which scores $51.0$ and  $50.6$.
These substantial performance improvements over prior methods highlight the effectiveness of our proposed Self-Selection-RGP method.
\item With LLama2-13B-Chat as the base LLM, our Self-Selection-RGP method consistently delivers strong performance on both TriviaQA and NQ datasets. 
In particular, Self-Selection-RGP exhibits superior performance on the TriviaQA dataset, achieving an accuracy of $66.0$ and $66.8$ under zero-shot and few-shot settings, respectively. 
These results reflect improvements of $2.2$ and $1.8$ over the best baseline method SURE, which scores $63.8$ and $65.0$ in accuracy.
In comparison, our Self-Selection-RGP demonstrates competitive performance on the NQ dataset relative to the SURE method.
In the few-shot setting, Self-Selection-RGP attains an accuracy of $51.6$, resulting in a $1.2$ improvement over SURE’s accuracy of $50.4$. 
However, in the zero-shot setting, our Self-Selection-RGP scores $46.2$, which is $5.8$ lower than SURE’s accuracy of $52.0$.
This performance disparity may be attributed to the fact that the questions in the NQ dataset are inherently more challenging for LLMs compared to those in TriviaQA, as evidenced by the performance of the LLM-only approach in \tabref{tab:main}.
This complexity hampers our method's ability to effectively distinguish the correct answers from the incorrect ones.
\item In comparison, the Self-Selection-Ori method demonstrates superior performance on the TriviaQA dataset in both zero-shot and few-shot settings when utilizing Mistral-7B as the base LLM, surpassing all baseline methods.
However, the Self-Selection-Ori method encounters difficulties in producing accurate results on the NQ dataset or when using LLama2-13B-Chat as the base LLM. 
The potential reasons for the performance disparity are twofold: 1) Mistral-7B exhibits more advanced reasoning capabilities compared to the LLama2-13B series models, as supported by the comparative analysis presented in the Mistral-7B technical report~\cite{jiang2023mistral7b}; and 2) the questions in the NQ dataset are relatively more challenging than those in TriviaQA for LLMs, as aforementioned.
\end{itemize}

\subsection{Impact of Different Retrieval Settings (RQ2)}
\subsubsection{\textbf{ Effect of Different Retriever}}
We experiment to demonstrate the compatibility of the proposed \framework~framework with different retrieval methods. 
Specifically, beyond the BGE retriever considered in Table \ref{tab:main}, we also use BM25 as the retriever, and compare our \approach~method with Standard RAG to analyze whether our method can still maintain superiority in performance over Standard RAG with the new retriever.
We adopt Mistral-7B and Llama2-13B-Chat as the base language models, respectively.
This results in the following four RAG systems with BM25 as their retriever: Standard RAG (7B) and Self-Selection-RGP (7B), where the base LLM is Mistral-7B, as well as Standard RAG (13B) and Self-Selection-RGP (13B), where the base LLM is Llama2-13B-Chat.
We present the performance of these four systems on both the NQ and TriviaQA datasets under zero-shot and few-shot settings in \Figref{fig:retriever}.
From the figure, we make the following observations: 

\begin{itemize}[leftmargin=*,nosep]
\item  With BM25 as the retriever, our proposed Self-Selection-RGP method consistently outperforms the Standard RAG method in each setting as illustrated in \Figref{fig:retriever}, revealing the compatibility and generalizability of our method regarding new retrieval techniques like BM25.
To be more specific, as shown in \Figref{fig:retriever} (a) and (c), in the zero-shot setting, our Self-Selection-RGP (7B) and Self-Selection-RGP (13B) outperform the Standard RAG (7B) and Standard RAG (13B) models by substantial margins on all three metrics over both NQ and TriviaQA datasets. 
This highlights the significant effectiveness of our Self-Selection-RGP method in the zero-shot scenarios.
Comparably, as shown in \Figref{fig:retriever} (b) and (d), in the few-shot settings, the performance gains achieved by our Self-Selection-RGP method are relatively small. 
This may be because the samples available in the few-shot scenarios can provide sufficient knowledge to LLMs, diminishing the relative advantage of our proposed approach compared to the more standard RAG method.

\item Our Self-Selection-RGP method demonstrates enhanced robustness and stability compared to the Standard RAG method. 
This is evident in the comparisons shown in \Figref{fig:retriever}, where the performance variation between zero-shot and few-shot settings for Self-Selection-RGP is significantly smaller than that observed for the Standard RAG method, as illustrated in both (a) versus (b) and (c) versus (d). 
This advantage stems from the Self-Selection-RGP's lower sensitivity to input noise or perturbations, achieved through its preference alignment training.
This makes Self-Selection-RGP more adaptable to diverse retrieval techniques, even when the retrieved passages are not precisely relevant. 
Therefore, the proposed Self-Selection-RGP is capable of producing more reliable and accurate results.
This improved resilience to input variation allows our Self-Selection-RGP to maintain more consistent performance across a wide range of task conditions, leading to its enhanced robustness and stability compared to the Standard RAG method. 
\end{itemize}

\begin{figure}[t]
\setlength\abovecaptionskip{-0.3px}
\setlength\belowcaptionskip{-8px}
  \centering
  \includegraphics[width=\linewidth]{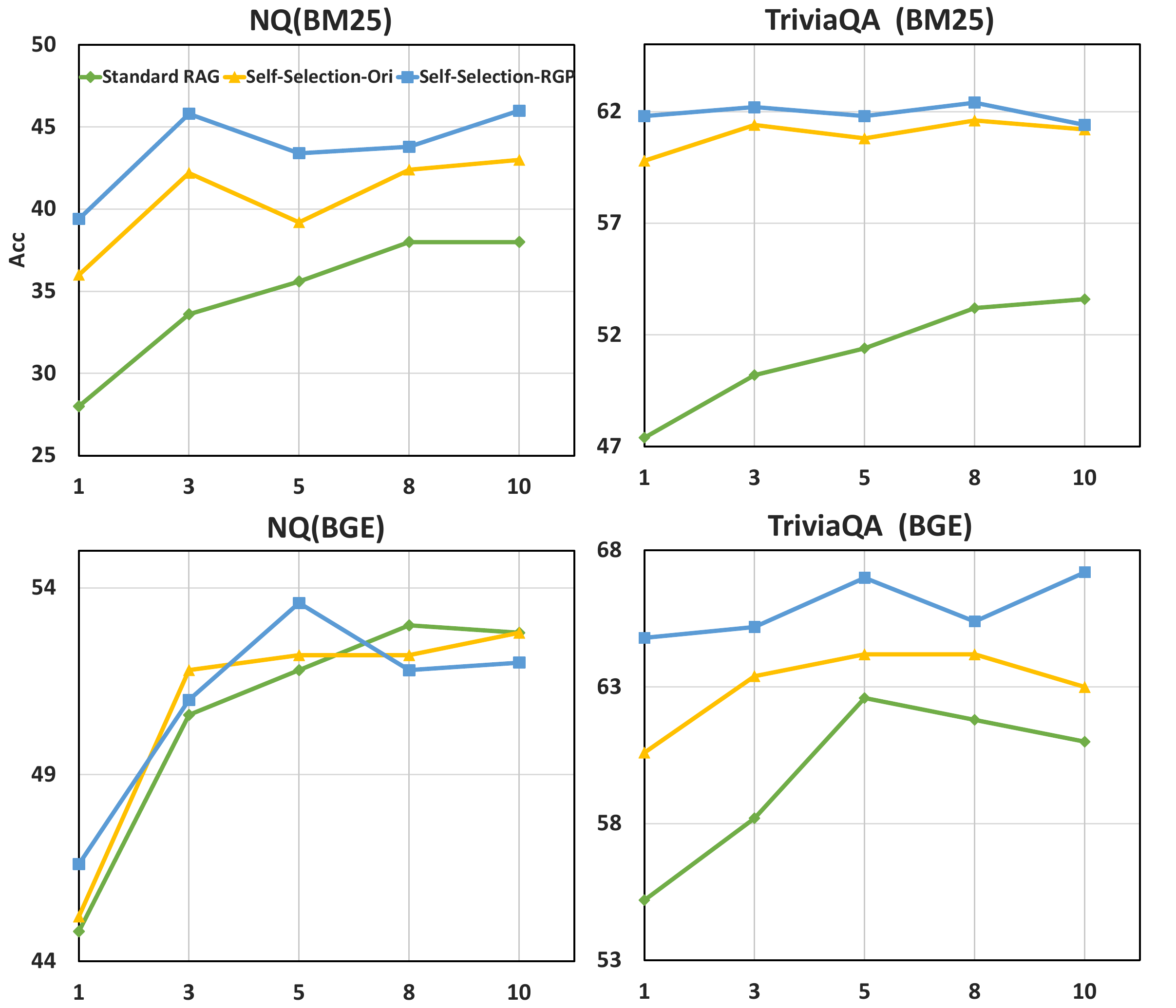}
  \caption{An illustration of the effects of varying the number of retrieved passages. } 
  \label{fig:num_passage}
  \vspace{-0.2cm}
\end{figure}

\subsubsection{\textbf{Effect of Number of Retrieved Passages}}
Next, we investigate the impact of the number of retrieved passages on the performance.
Simply increasing the number of retrieved passages by providing a broader range of external retrieval information has been one of the most straightforward and effective methods to improve the performance of retrieve-and-read systems~\cite{karpukhin-etal-2020-dense}. 
However, the effectiveness and applicability of this approach may be limited, as LLMs are constrained by the length of the input context~\cite{liu-etal-2024-lost}.
In the following, we evaluate the effects on our proposed methods by varying the number of retrieved passages, including $1$, $3$, $5$, $8$ and $10$.
Specifically, we adopt BGE and BM25 as the retriever and Mistral-7B as the base LLM in this experiment. 
We conduct a comparative analysis of performance across three RAG methods, i.e. Standard RAG, Self-Selection, and Self-Selection-RGP, on both NQ and TriviaQA datasets.
We summarize the experimental results in \Figref{fig:num_passage}.
From these results, we can make the following observations:
\begin{itemize}[leftmargin=*,nosep]
    \item Increasing the number of retrieved passages initially leads to noticeable improvements in the accuracy of all three RAG methods. However, once the number of retrieved passages exceeds a certain threshold, adding more passages has only marginal effects on the performance, and in some cases even degrades the overall performance.
    This may be due to information overload within the constrained context window of LLMs, which impairs the model’s capacity to accurately generate the correct responses.
    \item On the TriviaQA dataset, our Self-Selection-RGP method consistently yields the highest performance across varying numbers of retrieved passages and different retrievers. The Self-Selection method ranks second, while the Standard RAG method performs the worst.
    The performance comparison reveals the effectiveness of our proposed Self-Selection framework.
    \item On the NQ dataset, our performance of the three methods are similar to that on TriviaQA, with Self-Selection-RGP consistently ranking first across different numbers of retrieved passages.
    When the retriever BGE is adopted, all the three methods exhibit competitive performance across varying numbers of retrieved passages. 
    Note that Self-Selection-RGP method achieves the best performance when the number of retrieved passages is set to $5$
    Compared to the LLM-only approach, Standard RAG significantly enhances the performance by integrating retrieved passages into the LLM input, thereby establishing a strong baseline, as shown in \tabref{tab:main}. 
    Our method can still achieve comparable performance to this strong baseline, well showcasing its remarkable effectiveness across a varying number of retrieved passages.
\end{itemize}
 
\subsection{Analysis of Answer Generation Capability of LLMs (RQ3)}
\label{sec:llm-improve}

According to the experimental results presented above, the LLMs trained with the preference dataset have demonstrated remarkable improvements in distinguishing the correct responses from the incorrect ones.
One question naturally arises: ``To what extent does this training process affect the LLMs’ inherent ability in answer generation with or without the dependence on external retrieved knowledge?''
To answer this question, we design an experiment to compare LLMs' answer generation performance before and after preference alignment training. 
We adopt Mistral-7B as the base LLM in this analysis.
We first apply \approach~to train Mistral-7B using the augmented RGP dataset, resulting in a trained LLM that is referred to as Self-Selection-RGP-7B.

\begin{figure}[t]
\setlength\abovecaptionskip{-0.4px}
\setlength\belowcaptionskip{-4px}
  \centering
  \includegraphics[width=\linewidth]{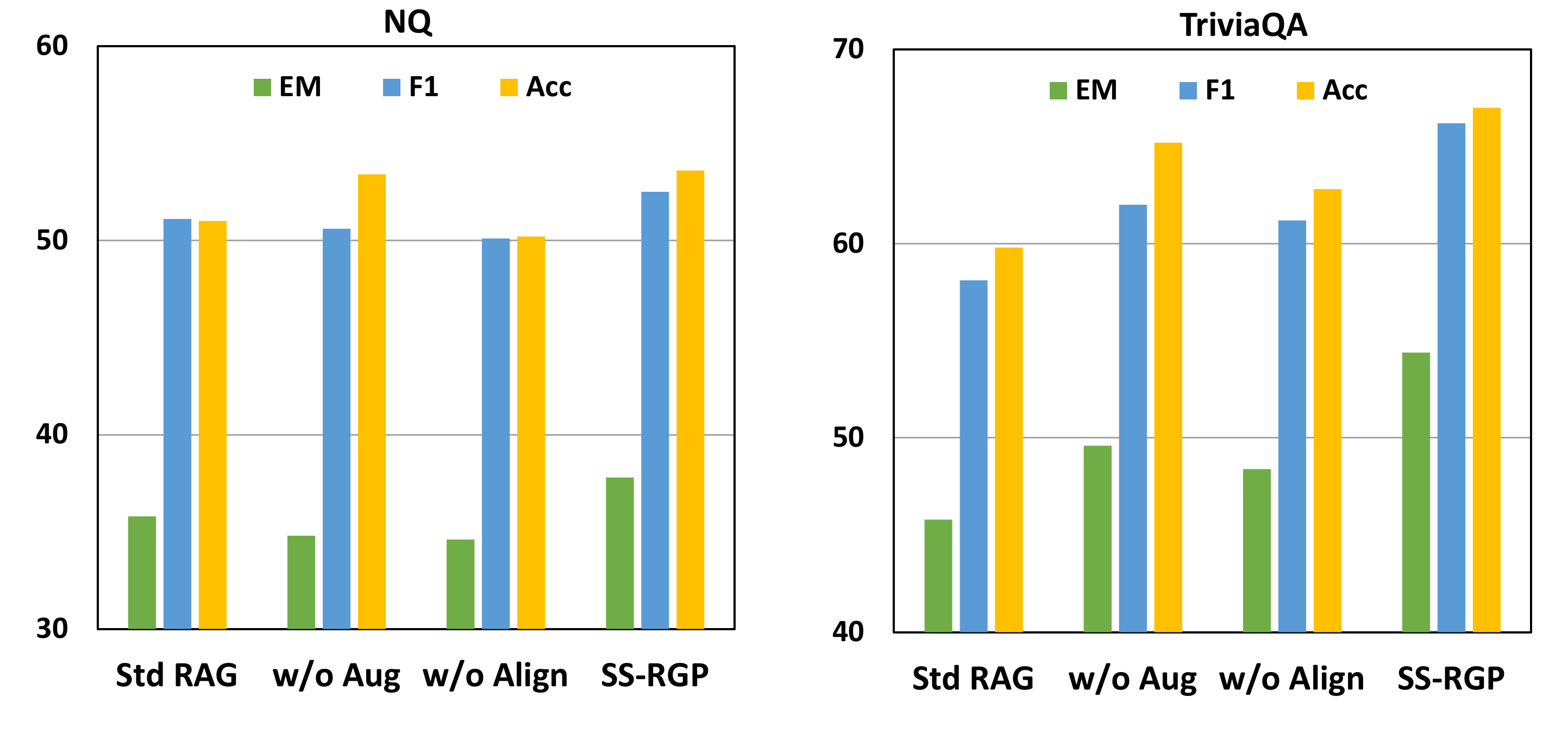}
  \caption{Ablation study. ``Std RAG'' refers to Standard RAG; ``w/o Aug'' indicates the method without Dataset Augmentation; ``w/o Align'' denotes the method without Preference Alignment; and ``SS-RGP'' represents our proposed Self-Selection RAG method.} 
  \label{fig:ablation}
  \vspace{-0.4cm}
\end{figure}

\begin{figure*}[t]
\setlength\abovecaptionskip{-0.3px}
\setlength\belowcaptionskip{-0.4px}
  \centering
  \includegraphics[width=\linewidth]{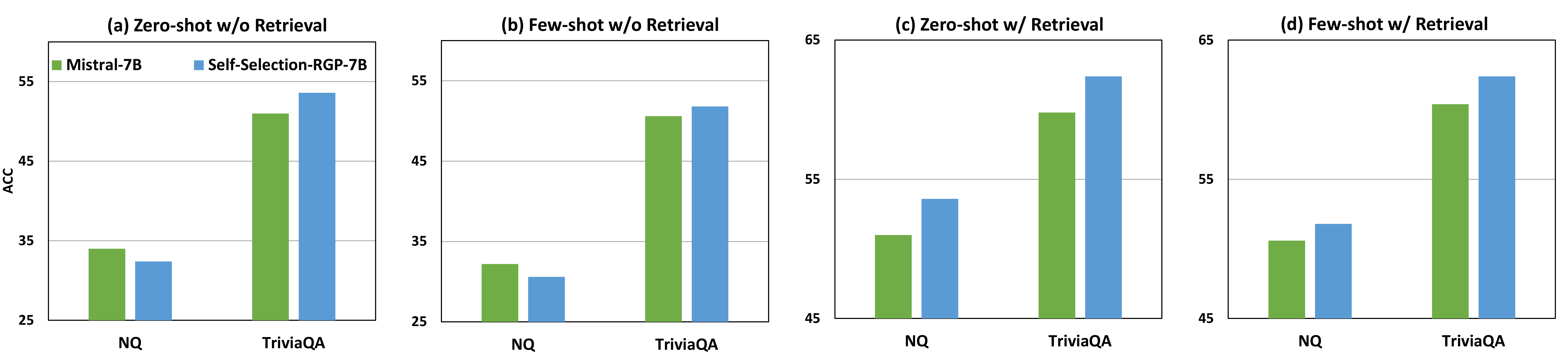}
  \caption{Analysis of the answer generation capability before and after preference alignment training. ``Mistral-7B'' denotes the vanilla LLM before preference alignment training; ``Self-Selection-RGP-7B'' refers to the LLM obtained after training Mistral-7B with the preference alignment dataset.} 
  \label{fig:llm-ans}
\end{figure*}

To comprehensively compare the answer generation capabilities of \emph{Mistral-7B} and \emph{Self-Selection-RGP-7B}, we conduct extensive experiments on NQ and TriviaQA datasets, under both zero-shot and few-shot settings, with and without the use of external knowledge.
We present the experimental results in \Figref{fig:llm-ans}, from which we make the following observations:
\begin{itemize} [leftmargin=*,nosep]
    \item As shown \Figref{fig:llm-ans} (a) and (b), without depending on external knowledge, the Self-Selection-RGP-7B exhibits improved capabilities in answer generation on TriviaQA while showing slightly worse performance on NQ, compared to the Mistral-7B, in both zero-shot and few-shot settings.  
    \item From \Figref{fig:llm-ans} (c) and (d), the Self-Selection-RGP-7B consistently outperforms Mistral-7B on both datasets when involving the retrieved relevant passages in their inputs, showing its enhanced answer generation abilities.
\end{itemize}
 
 The empirical findings presented above demonstrate that the Self-Selection-RGP-7B model exhibits enhanced capabilities not only in answer selection, but also in the broader task of answer generation.
 This suggests that the proposed Self-Selection-RGP method which trains LLMs on the augmented RGP dataset has led to notable improvements in LLMs' ability to generate high-quality answers, highlighting its potential for enhancing the overall response generation performance of LLMs.

\subsection{Ablation Study (RQ4)}

To verify the effect of each design in our proposed method, we conduct an ablation study with Mistral-7B on both NQ and TriviaQA datasets.
We utilize three methods for ablation experiments in addition to our proposed \approach~method, including 
1) \textbf{Standard RAG}, which appends the retrieved passages to the input of the vanilla LLM;
2) \textbf{w/o Dataset Augmentation}, which removes the step of dataset augmentation from our proposed method and trains Mistral-7B with the original RGP dataset only; 
3)\textbf{w/o Preference Alignment}, which removes the step of preference alignment for the LLM and applies the \framework~on the vanilla LLM.
All the methods are evaluated under the zero-shot setting for a fair comparison.
We present the results in \Figref{fig:ablation}, from which we make the following observations: 

\begin{itemize}[leftmargin=*,nosep]
\item Removing either the step of dataset augmentation or preference alignment results in performance drop in all evaluation metrics on both NQ and TriviaQA datasets, highlighting the rationale of the design in our proposed \approach~method. 
\item Comparably, the removal of preference alignment results in a more substantial decrease in performance, revealing the importance of teaching LLMs how to choose the correct answer from multiple candidates.
\item Compared to the Standard RAG method, our method consistently yields superior performance on both datasets, demonstrating its great effectiveness in enhancing RAG systems. 

\end{itemize}

\subsection{Error Analysis}
We conduct an error analysis to investigate the limitations of our \approach~method. 
With the Mistral-7B as the base LLM, we sample $100$ from the errors made by our \approach~method on the TriviaQA dataset to conduct our analysis. 
We categorize the errors into five groups, as shown in ~\tabref{tab:error-case}, each with an example:
(1) Lack of Evidence (51\%): The LLM itself does not have sufficient internal knowledge to answer the question, and the retrieved passages also fail to provide enough information;
(2) Partial Matching (20\%): The final prediction captures part of the correct answer only;
(3) Reasoning Error (14\%): The generated explanation(s) contain the answer or the relevant information to infer the answer, but the LLM fails to predict the answer;
(4) Selection Error(12\%): One of the pairwise predictions is correct,  but the model fails to identify the correct one;
(5) Formatting Error(3\%): The correct answer is included in the prediction but the output format does not follow the instruction, leading to a failed interpretation.

From the table, we make the following observations:
\begin{itemize}
    \item  Over half of the errors are caused by the failure to obtain the knowledge that is necessary for predicting the answer, i.e., Lack of Evidence. This reflects the importance of developing advanced techniques to fetch the relevant information given a question from external knowledge bases to complement LLMs' internal knowledge for producing more accurate results.
    \item Approximately $39\%$ of errors, including Partial Matching, Reasoning Errors, and Formatting Errors, originate from LLMs' inadequacies in accurately interpreting the human instruction or the relevant knowledge required to infer the precise answer in the correct format. To address this issue, LLMs with enhanced reasoning capabilities are required.  
    \item $12\%$ errors arise from the LLMs' inability to effectively distinguish the correct answer from plausible ones, underscoring the demand for LLMs with enhanced reasoning capabilities. 
\end{itemize}

\newcolumntype{L}[1]{>{\raggedright\arraybackslash}p{#1}}

\begin{table*}[ht]
\setlength\abovecaptionskip{-0.3px}
\setlength\belowcaptionskip{-0.1px}
\caption{Examples of errors and corresponding percentages. 
Q, G, and P denote the question, golden answer, and prediction, respectively. C1 and C2 represent two candidate responses.}
\label{tab:error-case} 
\centering
\renewcommand{\arraystretch}{1.2} 
\setlength{\tabcolsep}{4pt} 
\footnotesize
\begin{tabular}{L{0.1\textwidth}|L{0.85\textwidth}}
\toprule
\multirow{6}{*}{\makecell[l]{Lack of Evidence \\ (51\%)}} 
& Q: The Express, The Telegraph, The Economist, The Times, The Star, and The Independent were all what in 1830's London? (Not Newspapers) \\
& G: \textcolor{blue}{Stagecoaches} \\
& C1: Explanation: The Express, The Telegraph...were not entities in London during the 1830s... Answer: \textcolor{red}{Unknown}\\
& C2: Explanation: ...the passages provided only discuss their histories and developments starting from the mid-1800s... Answer: \textcolor{red}{Unknown} \\
& P: Explanation: The Express, The Telegraph...were not entities in London during the 1830s... Answer: \textcolor{red}{Unknown} \\
\midrule

\multirow{6}{*}{\makecell[l]{Partial Matching \\ (20\%)}} 
& Q: In which Parliamentary constituency did Michael Portillo replace Alan Clark as MP in 1999? \\
& G: \textcolor{blue}{Kensington and Chelsea (borough)} \\
& C1: Explanation: ...Portillo is most famously known for defeating Clark...in the constituency of Wellingborough. Answer: \textcolor{red}{Wellingborough}.\\
& C2: Explanation: Michael Portillo replaced Alan Clark as MP in Kensington and Chelsea... Answer: \textcolor{blue}{Kensington and Chelsea}\textcolor{gray}{ (borough)} \\
& P: Explanation: Michael Portillo replaced Alan Clark as MP in Kensington and Chelsea... Answer: \textcolor{blue}{Kensington and Chelsea}\textcolor{gray}{ (borough)} \\
\midrule

\multirow{7}{*}{\makecell[l]{Reasoning Error \\ (14\%)}} 
& Q: Richard Nixon was Vice President to which US state? \\
& G: \textcolor{blue}{Dwight D Eisenhower} \\
& C1: Explanation: Richard Nixon served as the Vice President under \textcolor{blue}{Dwight D. Eisenhower} from 1953 to 1961... However, it's important to note that Nixon didn't directly represent Kansas as Vice President; instead, he assisted President Eisenhower. Answer: \textcolor{red}{Kansas (through Eisenhower)} \\
& C2: Explanation: Richard Nixon was the Vice President of the United States for the state of California. Answer: \textcolor{red}{California.}\\
& P: Explanation: Richard Nixon was the Vice President of the United States for the state of California. Answer: \textcolor{red}{California.} \\
\midrule

\multirow{6}{*}{\makecell[l]{Selection Error \\ (12\%)}} 
& Q: The Greek word praxis refers to what, contrasting with theory? \\
& G: \textcolor{blue}{Practice} \\
& C1: Explanation: In ancient Greek philosophy, the term "praxis" refers to the application or practice of... Answer: \textcolor{blue}{practice} \\
& C2: Explanation: The Greek word praxis refers to the process of enacting... Answer: \textcolor{red}{The enactment or realization of a theory} \\
& P: Explanation: The Greek word praxis refers to the process of enacting... Answer: \textcolor{red}{The enactment or realization of a theory} \\
\midrule

\multirow{6}{*}{\makecell[l]{Formatting Error \\ (3\%)}} 
& Q: What do 'Barb' charts show? \\
& G: \textcolor{blue}{Television Ratings} \\
& C1: Explanation: Bar charts are a type of chart that display... Answer: \textcolor{red}{Bar charts show categorical data...} \\
& C2: Explanation: ...and other research findings related to television audience measurement and ratings in the United Kingdom. Answer: \textcolor{red}{Barb charts show comparisons}... In the context of \textcolor{blue}{television ratings}... \\
& P: Research ... television audience measurement and ratings in the United Kingdom. In the context of \textcolor{blue}{television ratings}... \\
\bottomrule
\end{tabular}
\end{table*}

\section{RELATED WORK}

\subsection{Retrieval-Augmented Generation}
Retrieval-Augmented Generation (RAG) \cite{Lewis2020RAG,Guu2020REALM} has been widely used for improving the performance of LLMs across various tasks by incorporating an Information Retriever (IR) module to leverage external knowledge. 
Most RAG systems~\cite{Lewis2020RAG,ram2023context,Izacard2023Atlas} integrate retrieved knowledge (e.g., passages) directly into the input, where the LLMs generate answers based on the external information obtained with the IR module.
There are also some methods utilizing Chain of Thought (CoT) \cite{wei2022cot, trivedi2023interleaving} or task decomposition \cite{Xu2024Search, wang2024selfdc, kim2024sure} to integrate external knowledge in intermediate reasoning steps or sub-tasks. 
Though effective, such indiscriminate use of external knowledge may introduce noise, degrading the quality of generated responses.
To address this problem, conditional use of external knowledge in RAG has been investigated.
Some works decide whether to retrieve and utilize external knowledge based on query characteristics, such as assessing query complexity through entity frequency \cite{mallen2023trust}, searching for similar questions \cite{wang2023skr}, or collecting question sets for training \cite{jeong2024adaptive}.
Some other works determine whether to integrate external knowledge according to the next token generation probability by LLMs, 
such as Self-RAG~\cite{asai2024selfrag}, FLARE\cite{jiang2023active}, DRAGIN~\cite{su2024dragin}, and Self-DC~\cite{wang2024selfdc}. 
In addition to these adaptive retrieval approaches, relevance-based methods~\cite{zhang2023merging,Xu2024Search,liu2024raisf} employ a relevance verification module to filter retrieved passages. 
For example, RA-ISF~\cite{liu2024raisf} assesses the relevance of retrieved passages by training a small LLM.
However, adaptive retrieval methods often rely solely on the input query or generated tokens, which limits their effectiveness as they may only acquire incomplete information. 
In comparison, relevance-based methods heavily rely on an additional verification module, leading to increased complexity of RAG systems and high sensitivity of the final response's quality to its verification accuracy. 
In this work, we propose a novel approach that leverages the LLM itself to holistically evaluate and reconcile responses from only its internal parametric knowledge and also from externally retrieved information, aiming to deliver more accurate responses.

\subsection{Preference Alignment for LLMs}
Preference alignment has emerged as an effective approach for improving the reliability of LLMs~\cite{Ouyang2022Training} by enabling them to evolve from their generated responses and environmental feedback.
Among existing preference alignment techniques, Reinforcement Learning from Human Feedback (RLHF) leverages human-provided feedback to train reward models, ensuring that LLMs produce responses aligned well with human preferences~\cite{Christiano2017RLHF,ziegler2019finetuning}.
RLHF has been shown to improve both the performance and the user-friendliness of LLMs in various Natural Language Processing (NLP) tasks, including summarization \cite{Stiennon2020summarize}, question answering \cite{Nakano2021WebGPT}, and instruction following \cite{Ouyang2022Training}.
However, RLHF requires extensive human annotation to train the reward model and involves a complex three-stage process, resulting in limited scalability and high training complexity.
To improve the scalability in preference alignment, Reinforcement Learning from AI Feedback (RLAIF)~\cite{Bai2022ConstitutionalAH,lee2024rlaif} utilizes the feedback from the LLM itself to train a reward model to optimize LLM performance through reinforcement learning. 
Direct Preference Optimization (DPO) \cite{Rafailov2023DPO} defines preference loss directly via a change of variables, which treats the LLM itself as its reward model.
By eliminating the need for an additional reward model, DPO substantially reduces the complexity involved in preference alignment training. 
In RAG systems, some studies utilize the signals generated by LLMs to optimize the retriever~\cite{Bonifacio2022InPars,shi2024replug} to retrieve LLM-preferred data, while other works align LLMs with specific domain knowledge and specific tasks through reinforcement learning~\cite{zhang2024knowledgeable,Yang2024IMRAG,Salemi2024Optimization,dong2024understandllmneedsdual,song2024measuring}.
In this work, we generate a preference dataset automatically and utilize it to strengthen LLMs' answer selection and generation capabilities in RAG systems via DPO.

\section{CONCLUSION}
In this work, we propose a novel \framework~framework to improve the accuracy and reliability of responses generated by LLMs in RAG systems.
Our method allows the LLM to select the more accurate one from a pair of responses generated based on internal parametric knowledge solely and by integrating external retrieved knowledge, to achieve enhanced performance. 
To strengthen the capabilities of the LLM in generating and selecting correct answers, we develop a \approach~method that trains the LLM with Direct Preference Optimization over a newly built Retrieval-Generation Preference (RGP) dataset. 
We conduct extensive experiments and analyses, which well validate the effectiveness of the proposed method. 
We hope this work paves the way for the development of more robust and reliable LLMs in RAG.


\bibliographystyle{ACM-Reference-Format}
\bibliography{sample-base}




\end{document}